\documentclass[runningheads]{llncs}
\usepackage{graphicx}
\usepackage{amsmath,amssymb} 
\usepackage{color}

\begin{document}
\pagestyle{headings}
\mainmatter

\def\ACCV18SubNumber{008}  

\title{Auto-Classification of Retinal Diseases in the Limit of Sparse Data Using a Two-Streams Machine Learning Model} 

\author{
C.-H. Huck Yang*\inst{1,2}\and
 Fangyu Liu* \inst{3}\and Jia-Hong Huang*\inst{1}\and
Meng Tian\inst{4}\and
Hiromasa Morikawa\inst{1}\and I-Hung Lin M.D. \inst{5}\and Yi Chieh Liu\inst{6}\and Hao-Hsiang Yang\inst{6}\and
Jesper Tegn\`er\inst{1,7}}


\authorrunning{C.-H. Huck Yang et al.}

\institute{
Biological and Environmental Sciences and Engineering Division,
Computer, Electrical and Mathematical Sciences and Engineering
Division, King Abdullah University of Science and Technology, Thuwal, Saudi Arabia \email{\{chao-han.yang, jiahong.huang, hiromasa.morikawa, jesper.tegner\}}@kaust.edu.sa\\ \and
Georgia Institute of Technology, GA, USA\\\and 
University of Waterloo, Canada \email{fangyu.liu@uwaterloo.ca}\\\and 
Department of Ophthalmology, Bern University Hospital, Bern, Switzerland  \and Department of Ophthalmology, Tri-Service General Hospital, Taiwan\and National Taiwan University, Taiwan \and Unit of Computational Medicine, Center for Molecular Medicine,
Department of Medicine, Karolinska Institutet, Sweden}

\maketitle
\linespread{0.9}

\begin{abstract}
Automatic clinical diagnosis of retinal diseases has emerged as a promising approach to facilitate discovery in areas with limited access to specialists. Based on the fact that fundus structure and vascular disorders are the main characteristics of retinal diseases, we propose a novel visual-assisted diagnosis hybrid model mixing the support vector machine (SVM) and deep neural networks (DNNs). Furthermore, we present a new clinical retina labels collection sorted by the professional ophthalmologist from the educational project Retina Image Bank, called EyeNet, for ophthalmology incorporating 52 retina diseases classes. Using EyeNet, our model achieves 90.40\% diagnosis accuracy, and the model performance is comparable to the professional ophthalmologists.
\end{abstract}

\section{Introduction}

Computational retinal disease methods \cite{tan2009detection,lalezary2006baseline} has been investigated extensively through different signal processing techniques. Retinal diseases are accessible to machine learning techniques due to their visual nature in contrast to other common human diseases requiring invasive techniques for diagnosis or treatments. Typically, the diagnosis accuracy of retinal diseases based on the clinical retinal images is highly dependent on the practical experience of a physician or ophthalmologist. However, training highly-skilled ophthalmologists usually take years and the number of them, especially in the less-developed area, is still far from enough. Therefore, developing an automatic retinal diseases detection system is important, and it will broadly facilitate diagnostic accuracy of retinal diseases. Moreover, for remote rural areas, where there are even no ophthalmologists locally to screen retinal disease, the automatic retinal diseases detection system also helps non-ophthalmologists find the patient of the retinal disease, and further, refer them to the medical center for further treatment.

\begin{figure}
\begin{center}
   \includegraphics[width=1.0\linewidth]{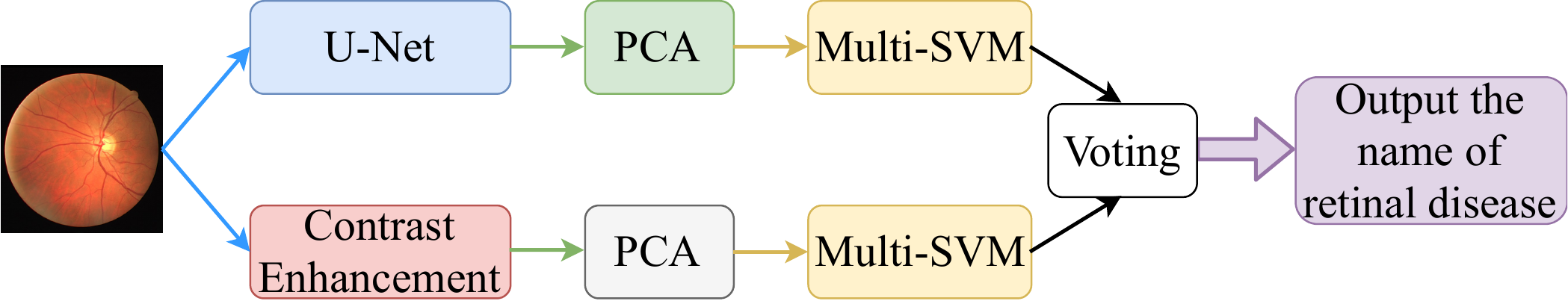}
\end{center}
   \caption{This figure represents our proposed two-streams model. A raw retinal image as an input of DNNs, U-Net, and as the other input to a contrast enhancement algorithm. Then we pass the output of U-Net to two separated PCA processing. Finally, the output of these two PCA modules is sent as inputs to the retina disease classifier, SVM, which give the outcome of predicted retina disease.}
\label{fig:figure1}
\end{figure}

The development of automatic diseases detection (ADD) \cite{sharifi2002classified} alleviates enormous pressure from social healthcare systems. Retinal symptom analysis \cite{abramoff2010retinal} is one of the important ADD applications given that it offers a unique opportunity to improve eye care on a global stage. The World Health Organization estimates that age-related macular degeneration (AMD) and Diabetic Retinopathy, which are two typical retinal diseases, are expected to affect over 500 million people worldwide by 2020 \cite{pizzarello2004vision}. 
  
Moreover, the increasing number of cases of diabetic retinopathy globally requires extending efforts in developing visual tools to assist in the analytic of the series of retinal disease. These decision support systems for retinal ADD, as \cite{bhattacharya2014watermarking} for non-proliferative diabetic retinopathy have been improved from recent machine learning success on the high dimensional images processing by featuring details on the blood vessel. \cite{lin2000rotation} demonstrated an automated technique for the segmentation of the blood vessels by tracking the center of the vessels on Kalman Filter. However, these pattern recognition based classification still rely on hand-crafted features and only specify for evaluating single retinal symptom. Despite extensive efforts using wavelet signal processing, retinal ADD remains a viable target for improved machine learning techniques applicable for point-of-care (POC) medical diagnosis and treatment in the aging society \cite{cochocki1993neural}. 

To the best of our knowledge, the amount of clinical retinal images are less compared to other cell imaging data, such as blood cell and a cancer cell. However, a vanilla deep learning based diseases diagnosis system requires large amounts of data. Therefore, we propose a novel visual-assisted diagnosis algorithm which is based on an integration of the support vector machine and deep neural networks. The primary goal of this work is to automatically classify 52 specific retinal diseases for human beings with the reliable clinical-assisted ability on the intelligent medicine approaches. To foster the long-term visual analytics research, we also present a visual clinical label collection, EyeNet, including several crucial symptoms as AMD, DR, uveitis, BRVO, BRAO.

\vspace{+0.3cm}
\noindent\textbf{Contributions.}
\begin{itemize}
    \item We design a novel two-streams-based algorithm on the support vector machine and deep neural networks to facilitate medical diagnosis of retinal diseases. 
    \item We present a new clinical labels collection, EyeNet, for Ophthalmology with 52 retina diseases classes as a crucial aid to the ophthalmologist and medical informatics community. 
    \item Finally, we visualize the learned features inside the DNNs model by heat maps. The visualization helps in understanding the medical comprehensibility inside our DNNs model.
\end{itemize}

\section{Related Work}
In this section, we review some works related to our proposed method. We divide the related works into three parts including medical dataset comparison, dimension reduction by feature extraction, and image segmentation by neural networks.

\noindent\textbf{2.1 Medical Dataset Comparison}

Large-scale datasets help the performance of deep learning algorithms comparable to human-level on the tasks of speech recognition \cite{hannun2014deep}, image classification and recognition \cite{deng2009imagenet}, and question answering \cite{rajpurkar2016squad,antol2015vqa,huang2017vqabq,huang2017novel,huang2017robustness}. In the medical community, large scale medical datasets also help algorithms achieve expert-level performance on detection of skin cancer \cite{esteva2017dermatologist}, diabetic retinopathy \cite{gulshan2016development}, heart arrhythmias \cite{rajpurkar2017cardiologist}, pneumonia \cite{rajpurkar2017chexnet}, brain hemorrhage \cite{grewal2018radnet}, lymph node metastases \cite{bejnordi2017diagnostic}, and hip fractures \cite{gale2017detecting}.  

Recently, the number of openly available medical datasets is growing. In Table \ref{table:table3}, we try to provide a summary of the publicly available medical image datasets related to ours. According to Table \ref{table:table3}, we notice that the recently released ChestXray14 \cite{wang2017chestx} is the largest medical dataset containing 112,120 frontal-view chest radiographs with up to 14 thoracic pathology labels. Moreover, the smallest medical dataset is DRIVE \cite{staal2004ridge} containing 40 retina images. Regarding the openly available musculoskeletal radiograph databases, the Stanford Program for Artificial Intelligence in Medicine and Imaging has a medical dataset containing pediatric hand radiographs annotated with skeletal age (AIMI). The Digital Hand Atlas \cite{gertych2007bone} includes the left-hand radiographs which are from children of various ages labeled with radiologist readings of bone age. Then, our proposed EyeNet contains 52 classes of diseases and 1747 images. 

\begin{table*}
\begin{center}
\scalebox{0.65}{
    \begin{tabular}{| l | l | l | l |}
    \hline
    ~~~\textbf{Name of Dataset} & \textbf{Study Type} & \textbf{Label} & \textbf{Number of Images}\\ \hline
    ~~\textbf{EyeNet} & ~~~~\textbf{Retina} & ~~~~~~\textbf{Labels mining of Retinal Diseases} & ~~~~~~\textbf{1747}\\ \hline
    ~~DRIVE \cite{staal2004ridge} & ~~~~Retina & ~~~~~~Retinal Vessel Segmentation  & ~~~~~~40\\ \hline
    ~~MURA \cite{rajpurkar2017mura} & ~~~~Musculoskeletal (Upper Extremity) & ~~~~~~Abnormality & ~~~~~~40,561\\ \hline
    Digital Hand Atlas \cite{gertych2007bone} & ~~~~Musculoskeletal (Left Hand) & ~~~~~~Bone Age & ~~~~~~1,390 \\  \hline   
    ChestX-ray14 \cite{wang2017chestx} & ~~~~Chest & ~~~~~~Multiple Pathologies & ~~~~~~112,120 \\  \hline
    DDSM \cite{heath2000digital}& ~~~~Mammogram & ~~~~~~Breast Cancer & ~~~~~~10,239 \\  \hline 
    \end{tabular}}
    \caption{Overview of available different types of medical label collection and image datasets.}
    \label{table:table3}
\end{center}
\end{table*}

\noindent\textbf{2.2 Dimension Reduction by Feature Extraction}

Feature extraction is a method to make the task of pattern classification or recognition easier. In image processing and pattern recognition, feature extraction is one of the special forms of dimensionality reduction \cite{costa2005classification} in some sense.
The purpose of feature extraction is to exploit the most relevant information based on the original data and describe the information in a space with lower dimensions. For example, typically the size of original medical image data, such as functional magnetic resonance imaging (fMRI) scans, is very large and it causes algorithms computationally inefficient. In this case, we will transform the original data into a reduced representation set of features. That is, we exploit a set of feature vectors to describe the original data and the process is called image feature extraction. In \cite{fukunaga2013introduction}, the authors mention that the representation by extracted feature vectors should have a dimensionality that corresponds to the intrinsic dimensionality of the original data. Then, the intrinsic dimensionality of data is the minimum number of parameters required to account for the properties of the original data. Moreover, the authors of \cite{jimenez1998supervised} claim that dimensionality reduction mitigates the curse of dimensionality and the other undesired properties of spaces with high dimensions. The dimensionality reduction by feature extraction method has been used in many different application fields such as document verification \cite{yang2016stacked}, character recognition \cite{trier1996feature}, extracting information from sentences \cite{huang2017novel,srihari1999information,huang2017vqabq}, machine translation \cite{somers1999example,bahdanau2014neural} and so on.

\noindent\textbf{2.3 Image Segmentation by Neural Networks}

Typically, researchers exploit the convolutional neural networks to do image classification tasks with a single class output label. However, in the biomedical image processing tasks, the output should contain localization. That is to say, a class label is assigned to each pixel. Furthermore, thousands of images in training set are typically beyond reach in the biomedical tasks. Therefore, the authors of \cite{ciresan2012deep} train a neural networks model, with sliding-window, to predict the output class label of each pixel by providing a sub-region, small patch, around that pixel as input. 

In \cite{ciresan2012deep}, we know that the neural network model can do localization and the number of training data, in the sense of patches, is much larger than the training images. Apparently, \cite{ciresan2012deep} has two drawbacks. First, there exists some trade-off between the use of context and localization accuracy. Then, since the model runs separately for each small patch and there is much redundancy due to overlapping patches, it is not efficient in the sense of computational speed. Recently, the authors of \cite{hariharan2015hypercolumns,seyedhosseini2013image} have proposed an approach which can do the good localization and use of context at the same time. 

In the U-Net paper \cite{ronneberger2015u}, the authors build upon an even more elegant neural network architecture, the so-called fully convolutional network \cite{long2015fully}. The authors modify the architecture such that it works with very few training images and produces even more accurate image segmentation. The main idea of \cite{long2015fully} is to supplement a usual contracting neural network by the successive layers. Then, the authors exploit upsampling operators to replace pooling operators, so the resolution of output is enhanced by these layers. In order to do localization, the authors combine the upsampled output and high-resolution features from the contracting path. Furthermore, a successive convolutional layer learns to assemble a more accurate output based on this information. Due to the advantages of U-Net mentioned above, we modify and incorporate the U-Net to our proposed method. 

\begin{figure*}
\begin{center}
   \includegraphics[width=1.0\linewidth]{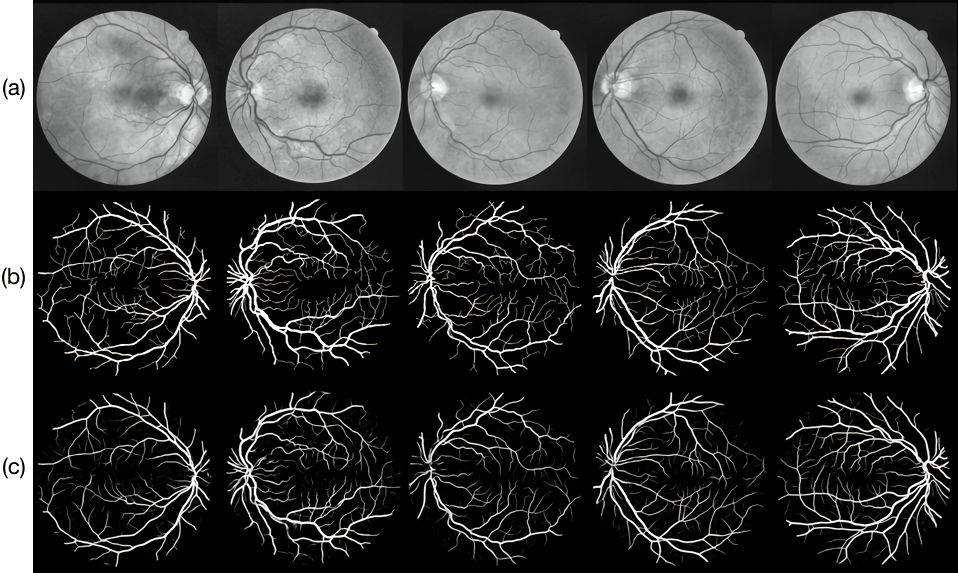}
\end{center}
   \caption{The figure shows the result of U-Net effects on (a),  unseen eyeball clinical images with different morphological shapes. (b) is the ground truth and (c) is the generated result of vessel-subtracted U-Net. Based on (b) and (c), we discover that the results are highly similar to the ground truth.}
\label{fig:figure2}
\end{figure*}

\section{Methodology}

In this section, we present the workflow of our proposed model, referring to Figure \ref{fig:figure1}.  

\noindent\textbf{3.1 U-Net}

DNNs has greatly boosted the performance of image classification due to its power of image feature learning \cite{simonyan2014very}. The active retinal disease is characterized by exudates around retinal vessels resulting in cuffing of the affected vessels \cite{khurana2007comprehensive}. However, ophthalmology images from clinical microscopy are often overlayed with white sheathing and minor features. Segmentation of retinal images has been investigated as a critical \cite{rezaee2017optimized} visual-aid technique for ophthalmologists. U-Net \cite{ronneberger2015u} is a functional DNNs, especially for segmentation. Here, we proposed a modified version of U-Net by reducing the copy and crop processes with a factor of two. The adjustment could speed up the training process and have been verified as an adequate semantic effect on small size images. We use cross-entropy for evaluating the training processes as:
\[
\ E = \sum_{x\in \Omega }w(x)log(p_{l}(x))   \hspace{+1.25cm} (1)
\]

where $p_{l}$ is the approximated maximum function, and the weight map is then computed as:
\[
\ w(x)=w_{c}(x)+w_{0}\cdot exp(\frac{-(d_{x1}+d_{x2})^2}{2\sigma^2}) \hspace{+0.5cm} (2)
\]

\begin{figure*}[t]
\begin{center}
   \includegraphics[width=1.0\linewidth]{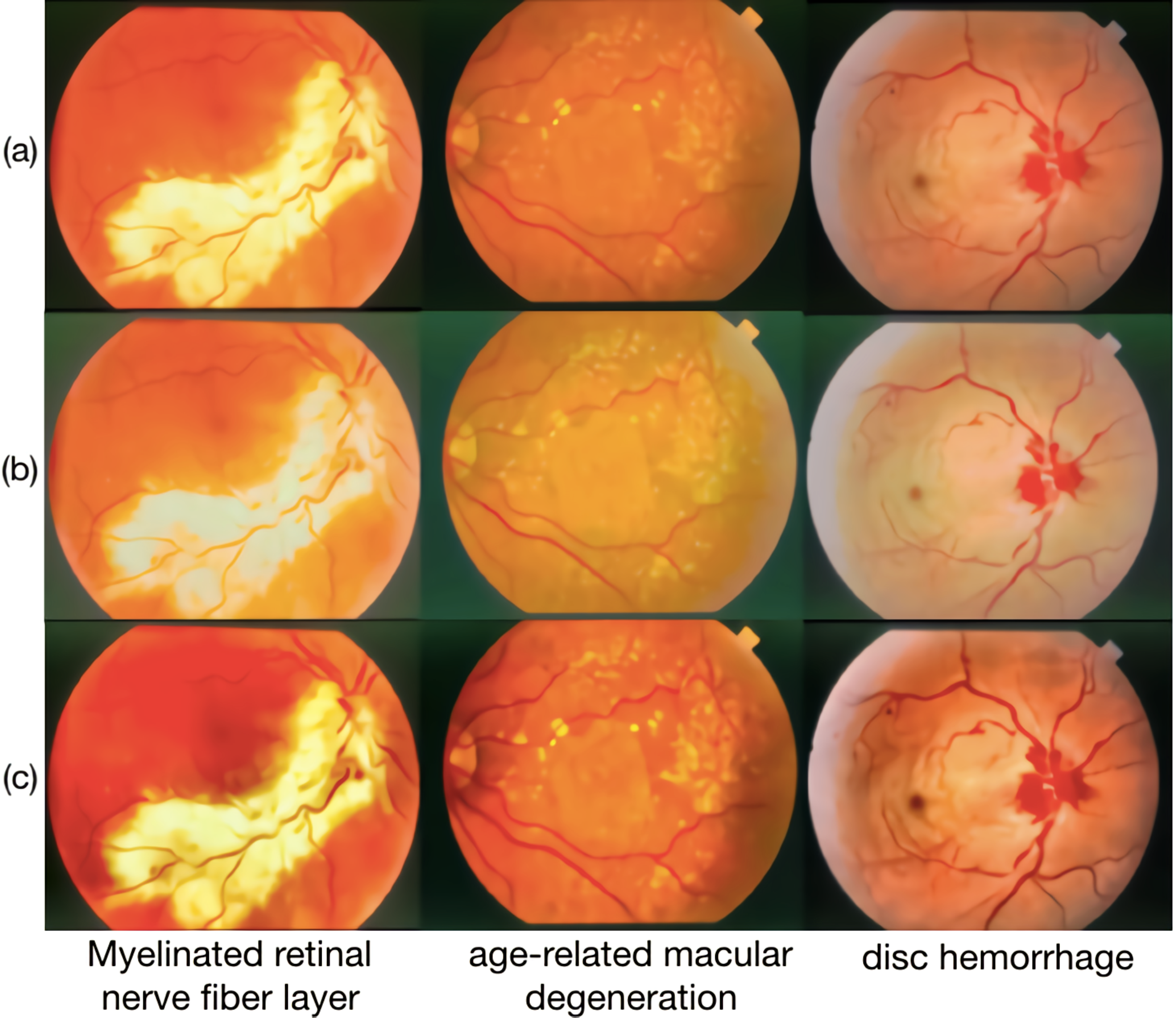}
\end{center}
   \caption{This figure illustrates the qualitative results of contrast enhancement algorithm from the original clinical images to the  (b) histogram equalization (c) contrast-limited adaptive histogram equalization.}
\label{fig:figure3}
\vspace{-0.3cm}
\end{figure*}

$d_{x1}$ designates the distance to the border of the nearest edges and $d_{x2}$ designates the distance to the border of the second nearest edges. LB score is shown as \cite{cochocki1993neural}. We use the deep convolutional neural network (CNN) of two $3\times3$ convolutions. Each step followed by a rectified linear unit (ReLU) and a $2\times2$ max pooling operation with stride 2 for downsampling; a layer with an even x- and y-size is selected for each operation. For the U-Net model, we use existing DRIVE \cite{staal2004ridge} dataset as the training segmentation mask. Then, we use Our proposed model converges at the 44th epoch when the error rate of the model is lower than $0.001$. The Jaccard similarity of our U-Net model is 95.59\% by validated on a 20\% test set among EyeNet shown in Figure \ref{fig:figure2}. This model is robust and feasible for different retinal symptoms as illustrated in Figure \ref{fig:figure3}. The area under ROC curve is $0.979033$ and the area under the Precision-Recall curve is $0.910335$. 

\noindent\textbf{3.2 Principal Component Analysis as Eigenface in the limit of Sparse Data}

\[
\lambda_{k}=\frac{1}{M}\sum_{n=1}^{M}(u^{T}_{k}\Phi_{n})^{2} ~~~~~(3)
\]

Eigenface \cite{turk1991face} is classical and high-efficient image recognition technique derived from the covariance matrix of the probability distribution over the high-dimensional vector space of face images. Even with a single training image, previous works \cite{lyons1999automatic,wu2002face} of eigenface already established robust automatic classification with confident accuracy ($85.6\%$) by combined principal component analysis (PCA) and SVM classifiers. As a biological feature, retinal images share similar properties with the human face for a potential with eigenface recognition \cite{moghaddam1998beyond} included finite semantic layout between facial features and ophthalmological features \cite{akram2011retinal}. The eigenface could be calculated \cite{turk1991face} by maximizing the equivalent $(3)$, where $\Phi_{n}$ represent the face differ, $u_{k}$ is a chosen $k_{th}$ vector, $\lambda_{k}$ is the $k_{th}$ eigenvalue, and $M$ is a number of the dimension space. 
In our experiment, we select the $k_{UNet} = 40$ and $k_{RGB}=61$ to generate a eigenface with highest accuracy for the U-Net-stream and RGB-stream separately.  

\noindent\textbf{3.3 Support Vector Machine}

Support Vector Machine is a machine learning technique for classification, regression, and other learning tasks. Support vector classification (SVC) in SVM, map data from an input space to a high-dimensional feature space, in which an optimal separating hyperplane that maximizes the boundary margin between the two classes is established.
The hinge loss function is shown as:
\[
\ \frac{1}{n}\left [ \sum_{i=1}^{n} max(0,1-y_{i}(\vec{w}\cdot\vec{x_{i}}-b))\right ]+\lambda \left \| \vec{w} \right \|^2 \hspace{+0.5cm} (4)
\]
Where the parameter $\lambda$ determines the trade off between increasing the margin-size and ensuring that the $\vec{x_{i}}$ lies on the right side of the margin. We use radial basis function (RBF) and polynomial kernel for SVC, which have been widely discussed \cite{kuo2014kernel} as a kernel-based fast SVC for images. 

\noindent\textbf{3.4 Contrast Enhancement}

Contrast enhancement techniques play a vital role in image processing to bring out the information that exists within a less dynamic range of that image. As a major clinical feature, fundus \cite{crick2003textbook,akram2005common} structure is highly related \cite{tang2015contrast} the image contrast \cite{noyel2017superimposition}. Here, we use histogram equalization for the contrast enhancement in retinal images. Compare to the original images, images after histogram equalization show the light color detail as lesions in Figure \ref{fig:figure3}(b). Images after contrast-limited adaptive histogram equalization (CLAHE) give further features as areas of retinopathy in Figure \ref{fig:figure3}(c).

\section{Efforts on Retinal Dataset}

Retina Image Bank (RIB) \cite{rib} is an international clinical project launched by American Society of Retina Specialists in 2012, which allow retina specialists and ophthalmic photographers around the world to share the existing clinical cases online for medicine-educational proposes for patients and physicians in developing countries lack training resource. Any researcher could join as a contributor for dedicating the retinal images or as a visitor using the medical images and label for non-commercial propose. With the recent success on dataset collection, such as ImageNet \cite{krizhevsky2012imagenet}, we believe that the effort of sorting and mining the clinical labels from RIB is valuable. With a more developer-friendly information pipeline, both Ophthalmology and Computer Vision community could go further on the analytical researches on medical informatics. Our proposed label collection, EyeNet is mainly based on the RIB and following the RIB's using guideline.

\begin{figure*}[t]
  \centering
\includegraphics[width=\linewidth]{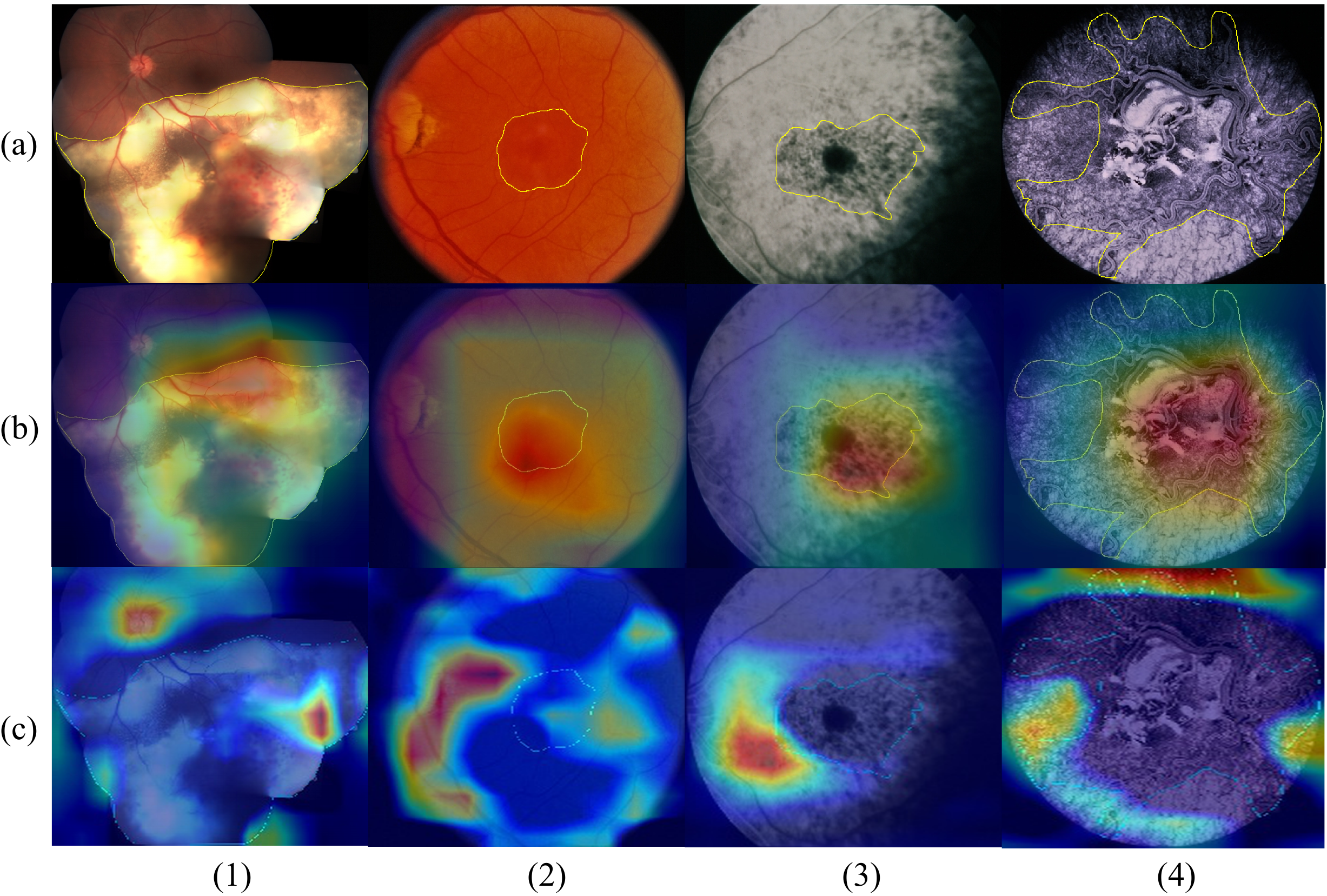}
    \caption{We use figure (i, j), where $i = a, b, c$ and $j = 1, 2, 3, 4$, to demonstrate that our proposed method can capture the similar lesion areas as the ophthalmologist's manual annotations, i.e., the yellow sketches.}
\label{fig:figure4}
\end{figure*}

\section{Experiments}

In this section, we describe the implementation details and experiments we conducted to validate our proposed method.

\noindent\textbf{5.1 Label Collection}

For experiments, the EyeNet is randomly divided into three parts: 70\% for training, 10\% for validation and 20\% for testing. All the training data have to go through the PCA before SVM. All classification experiments are trained and tested on the EyeNet.

\noindent\textbf{5.2 Setup}

The EyeNet has been processed to U-Net to generate a subset with a semantic feature of the blood vessel. For the DNNs and Transfer Learning models, we directly use the RGB images from the retinal dataset. EyeNet will be published online after getting accepted. For the CLAHE processing, we use $adapthisteq$ function from the image toolbox in MATLAB.

\noindent\textbf{5.3 Deep Convolutional Neural Networks}

CNN has demonstrated extraordinary performance in visual recognition tasks \cite{krizhevsky2012imagenet}, and the state of the art is in a great many vision-related benchmarks and challenges \cite{xie2017aggregated}. With little or no prior knowledge and human effort in feature design, it yet provides a general and effective method solving variant vision tasks in variant domains. This new development in computer vision has also shown great potential for helping/replacing human judgment in vision problems like medical imaging \cite{esteva2017dermatologist}, which is the topic we try to address in this paper. In this section, we introduce several baselines in multi-class image recognition and compare their results on the EyeNet.

\noindent\textbf{Baseline1-AlexNet}

AlexNet \cite{krizhevsky2012imagenet} brought up a succinct network architecture, with 3 fully connected layers, 5 convolutional layers, and the activation function is ReLU \cite{nair2010rectified}.

\noindent\textbf{Baseline2-VGG}

The authors of VGG \cite{simonyan2014very} exploit the filters (3x3) repeatedly to replace the large filters (5x5,7x7) in traditional architectures. By increasing depths of the network, it achieved better results on ImageNet with fewer parameters.

\noindent\textbf{Baseline3-ResNet}

Residual Networks \cite{hedeep}, one of the most popular neural networks today, utilize skip connections or short-cuts to jump over some layers. With skip connections, the network essentially collapses into a shallower network in the initial phase and this makes it easier to be trained, and then it expands its layers as it learns more of the feature space.

\noindent\textbf{Baseline4-SqueezeNet}

In real world, medical imaging tasks usually require a small and effective model to adapt to limited resources. As some deep neural networks cost several hundred megabytes to store, SqueezeNet \cite{iandola2016squeezenet} adopting model compression techniques has achieved the accuracy of AlexNet level with around 500 times smaller models.


\noindent\textbf{5.4 Transfer Learning}

We use a transfer learning framework from the normalized ImageNet \cite{krizhevsky2012imagenet} to the EyeNet for solving the small samples issue on the computational retinal visual analytics. With sufficient and utilizable training classified model, Transfer Learning resolves the challenge of Machine Learning in the limit of minimal amount of training labels and it drastically reduces the data requirements. The first few layers of DNNs learn features , similar to Gabor filters and color blobs, and these features appear not to be specific to any particular task or dataset and thus applicable to other datasets and tasks \cite{yosinski2014transferable}. Our experiments show the significant improvement after we apply the pretrained parameters on our deep learning based models, referring to Table \ref{table:table2} and Table \ref{table:table5}. 

\begin{table}[t]
\begin{center}
\scalebox{1.0}{
    \begin{tabular}{| l | l | l |}
    \hline
    \textbf{Hybrid-Ratio} & \textbf{RBF}& \textbf{Polyn.} \\ \hline
    ~~~0\%~:~100\%& ~~~~0.8159 & ~~~~~~~0.8391  \\ \hline
    ~~~40\%~:~60\%& ~~~~0.8371 & ~~~~~~~0.8381 \\ \hline
    ~~~50\%~:~50\% & ~~~~\textbf{0.9002} & ~~~~~~~0.8632 \\ \hline
    ~~~60\%~:~40\%& ~~~~0.8881 & ~~~~~~~\textbf{0.9040}  \\ \hline
    ~~~100\%~:~0\%& ~~~~0.8324 & ~~~~~~~0.8241  \\    
    \hline
    \end{tabular}}
    \caption{Accuracy comparison of the two-streams model with Radial basis function (RBF) and polynomial kernel. We use the hybrid-ratio \cite{yang2018novel} of the mixed weighted voting between two multi-SVCs trained from images over U-Net and CLAHE. }
    \vspace{-0.8cm}
\label{table:table1}
\end{center}
\end{table}

\begin{table}[t]
\begin{center}
\scalebox{1.0}{
    \begin{tabular}{| l | l | l |}
    \hline
    ~~~\textbf{Model} & \textbf{Pretrained} & \textbf{Random Init.} \\ \hline
    ~~AlexNet & ~~~~0.7912 & ~~~~~~0.4837  \\ \hline
    ~~VGG11 & ~~~~ 0.8802 & ~~~~~~\textbf{0.7579} \\ \hline
    ~~VGG13 & ~~~~0.8721 & ~~~~~~ 0.7123 \\ \hline
    ~~ResNet18 & ~~~~\textbf{0.8805} & ~~~~~~ 0.7250 \\ \hline
    SqueezeNet & ~~~~0.8239 & ~~~~~~0.5625  \\    
    \hline
    \end{tabular}}
    \caption{Accuracy comparison of three DNNs baselines on EyeNet.}
    \vspace{-0.3cm}
    \label{table:table2}
\end{center}
\vspace{-0.3cm}
\end{table}

\begin{table}[t]
\begin{center}
\scalebox{1.0}{
    \begin{tabular}{| l | l | l |}
    \hline
    ~~~\textbf{Model} & \textbf{Pretrained} & \textbf{Random Init.} \\ \hline
    ~~AlexNet & ~~~~0.7952 & ~~~~~~0.4892  \\ \hline
    ~~VGG11 & ~~~~ 0.8726 & ~~~~~~0.7583 \\ \hline
    ~~VGG13 & ~~~~\textbf{0.8885} & ~~~~~~ \textbf{0.7588} \\ \hline
    ~~ResNet18 & ~~~~0.8834 & ~~~~~~ 0.6741 \\ \hline
    SqueezeNet & ~~~~0.8349 & ~~~~~~0.5721  \\    
    \hline
    \end{tabular}}
    \caption{Accuracy comparison of three DNNs baselines on EyeNet.}
    \vspace{-0.8cm}
    \label{table:table5}
\end{center}
\end{table}

\noindent\textbf{5.5 Two-Streams Results}

\vspace{-0.1cm}
All SVM has implemented in Matlab with libsvm \cite{chang2011libsvm} module. We separate both the original retinal dataset and the subset to three parts included 70\% training set, 20\% test set, and 10\% validation set. By training two multiple-classes SVM models on both original EyeNet and the subset, we implement a weighted voting method to identify the candidate of retina symptom. We have testified different weight ratio as $Hybrid-Ratio$, SVM model with \{Images over CLAHE: Image over U-Net \}, with different accuracy at Table \ref{table:table1}. We have verified the model without over-fitting by the validation set via normalization on the accuracy with \~2.03\% difference.

\noindent\textbf{5.6 Deep Neural Networks Results}

All DNNs are implemented in PyTorch. We use identical hyperparameters for all models. The training lasts 400 epochs. The first 200 epochs take a learning rate of 1e-4 and the second 200 take 1e-5. Besides, we apply random data augmentation during training. In every epoch, there is $70\%$ probability for a training sample to be affinely transformed by one of the operations in \{flip, rotate, transpose\}$\times$\{random crop\}. Though ImageNet and our Retinal Dataset are much different, using weights pretrained on ImageNet rather than random ones has boosted test accuracy of any models with 5 to 15 percentages, referring to Table \ref{table:table2}. Besides, pretrained models tend to converge much faster than random initialized ones as suggested in Figure \ref{fig:figure4}. The performance of DNNs on our retinal dataset can greatly benefit from the knowledge of other domains.

\noindent\textbf{5.7 Neuron Visualization for Medical Images} 

Importantly, we verified the hypothesis that vessel-based segmentation and contrast enhancement are two coherent features to decide the type of retinal diseases. Using techniques of generating class activation maps introduced by \cite{zhou2015cnnlocalization}, we visualized feature maps of the final convolution layer of ResNet18 (which is one of our deep learning model baselines). We notice that the features learned by deep learning models agree with our intuitions about developing the two-stream machine learning model. In fact, in the clinical diagnosis process, "vessel patterns" and "fundus structure" are also the two most crucial features to identify the symptom of different diseases. These two types of features actually cover more than 80\% of retinal diseases \cite{crick2003textbook,akram2005common}.


\section{Conclusion and Future Work}

In this work, we have designed a novel hybrid model for visual-assisted diagnosis based on the SVM and U-Net. The performance of this model shows the higher accuracy, 90.40\%, over the other pre-trained DNNs models as an aid for ophthalmologists. Also, we propose the EyeNet to benefit the medical informatics research community. Finally, since our dataset not only contains images but also text information of the images, image captioning and Visual Question Answering \cite{huang2017vqabq,huang2017novel,huang2017robustness} based on the retinal images are also the interesting future directions. Our work may also help the remote rural area, where there are no ophthalmologists locally, to screen retinal disease without the help of ophthalmologists in the future.

\bibliographystyle{splncs}
\bibliography{egbib}

\begin{thebibliography}{10}

\bibitem{tan2009detection}
Tan, O., Chopra, V., Lu, A.T.H., Schuman, J.S., Ishikawa, H., Wollstein, G.,
  Varma, R., Huang, D.:
\newblock Detection of macular ganglion cell loss in glaucoma by fourier-domain
  optical coherence tomography.
\newblock Ophthalmology \textbf{116} (2009)  2305--2314

\bibitem{lalezary2006baseline}
Lalezary, M., Medeiros, F.A., Weinreb, R.N., Bowd, C., Sample, P.A., Tavares,
  I.M., Tafreshi, A., Zangwill, L.M.:
\newblock Baseline optical coherence tomography predicts the development of
  glaucomatous change in glaucoma suspects.
\newblock American journal of ophthalmology \textbf{142} (2006)  576--582

\bibitem{sharifi2002classified}
Sharifi, M., Fathy, M., Mahmoudi, M.T.:
\newblock A classified and comparative study of edge detection algorithms.
\newblock In: Information Technology: Coding and Computing, 2002. Proceedings.
  International Conference on, IEEE (2002)  117--120

\bibitem{abramoff2010retinal}
Abr{\`a}moff, M.D., Garvin, M.K., Sonka, M.:
\newblock Retinal imaging and image analysis.
\newblock Reviews in biomedical engineering \textbf{3} (2010)  169--208

\bibitem{pizzarello2004vision}
Pizzarello, L., Abiose, A., Ffytche, T., Duerksen, R., Thulasiraj, R., Taylor,
  H., Faal, H., Rao, G., Kocur, I., Resnikoff, S.:
\newblock Vision 2020: The right to sight: a global initiative to eliminate
  avoidable blindness.
\newblock Archives of ophthalmology \textbf{122} (2004)  615--620

\bibitem{bhattacharya2014watermarking}
Bhattacharya, S.:
\newblock Watermarking digital images using fuzzy matrix compositions and
  ($\alpha$, $\beta$)-cut of fuzzy set.
\newblock International Journal of Advanced Computing (2014)

\bibitem{lin2000rotation}
Lin, C.Y., Wu, M., Bloom, J.A., Cox, I.J., Miller, M.L., Lui, Y.M.:
\newblock Rotation-, scale-, and translation-resilient public watermarking for
  images.
\newblock In: Security and watermarking of multimedia contents II. Volume
  3971., International Society for Optics and Photonics (2000)  90--99

\bibitem{cochocki1993neural}
Cochocki, A., Unbehauen, R.:
\newblock Neural networks for optimization and signal processing.
\newblock John Wiley \& Sons, Inc. (1993)

\bibitem{hannun2014deep}
Hannun, A., Case, C., Casper, J., Catanzaro, B., Diamos, G., Elsen, E.,
  Prenger, R., Satheesh, S., Sengupta, S., Coates, A.,  et~al.:
\newblock Deep speech: Scaling up end-to-end speech recognition.
\newblock arXiv preprint arXiv:1412.5567 (2014)

\bibitem{deng2009imagenet}
Deng, J., Dong, W., Socher, R., Li, L.J., Li, K., Fei-Fei, L.:
\newblock Imagenet: A large-scale hierarchical image database.
\newblock In: Computer Vision and Pattern Recognition, 2009. CVPR 2009. IEEE
  Conference on, IEEE (2009)  248--255

\bibitem{rajpurkar2016squad}
Rajpurkar, P., Zhang, J., Lopyrev, K., Liang, P.:
\newblock Squad: 100,000+ questions for machine comprehension of text.
\newblock arXiv preprint arXiv:1606.05250 (2016)

\bibitem{antol2015vqa}
Antol, S., Agrawal, A., Lu, J., Mitchell, M., Batra, D., Lawrence~Zitnick, C.,
  Parikh, D.:
\newblock Vqa: Visual question answering.
\newblock In: Proceedings of the ICCV. (2015)  2425--2433

\bibitem{huang2017vqabq}
Huang, J.H., Alfadly, M., Ghanem, B.:
\newblock Vqabq: visual question answering by basic questions.
\newblock arXiv:1703.06492 (2017)

\bibitem{huang2017novel}
Huang, J.H., Dao, C.D., Alfadly, M., Ghanem, B.:
\newblock A novel framework for robustness analysis of visual qa models.
\newblock arXiv:1711.06232 (2017)

\bibitem{huang2017robustness}
Huang, J.H., Alfadly, M., Ghanem, B.:
\newblock Robustness analysis of visual qa models by basic questions.
\newblock arXiv:1709.04625 (2017)

\bibitem{esteva2017dermatologist}
Esteva, A., Kuprel, B., Novoa, R.A., Ko, J., Swetter, S.M., Blau, H.M., Thrun,
  S.:
\newblock Dermatologist-level classification of skin cancer with deep neural
  networks.
\newblock Nature \textbf{542} (2017)  115

\bibitem{gulshan2016development}
Gulshan, V., Peng, L., Coram, M., Stumpe, M.C., Wu, D., Narayanaswamy, A.,
  Venugopalan, S., Widner, K., Madams, T., Cuadros, J.,  et~al.:
\newblock Development and validation of a deep learning algorithm for detection
  of diabetic retinopathy in retinal fundus photographs.
\newblock Jama \textbf{316} (2016)  2402--2410

\bibitem{rajpurkar2017cardiologist}
Rajpurkar, P., Hannun, A.Y., Haghpanahi, M., Bourn, C., Ng, A.Y.:
\newblock Cardiologist-level arrhythmia detection with convolutional neural
  networks.
\newblock arXiv preprint arXiv:1707.01836 (2017)

\bibitem{rajpurkar2017chexnet}
Rajpurkar, P., Irvin, J., Zhu, K., Yang, B., Mehta, H., Duan, T., Ding, D.,
  Bagul, A., Langlotz, C., Shpanskaya, K.,  et~al.:
\newblock Chexnet: Radiologist-level pneumonia detection on chest x-rays with
  deep learning.
\newblock arXiv preprint arXiv:1711.05225 (2017)

\bibitem{grewal2018radnet}
Grewal, M., Srivastava, M.M., Kumar, P., Varadarajan, S.:
\newblock Radnet: Radiologist level accuracy using deep learning for hemorrhage
  detection in ct scans.
\newblock In: Biomedical Imaging (ISBI 2018), 2018 IEEE 15th International
  Symposium on, IEEE (2018)  281--284

\bibitem{bejnordi2017diagnostic}
Bejnordi, B.E., Veta, M., van Diest, P.J., van Ginneken, B., Karssemeijer, N.,
  Litjens, G., van~der Laak, J.A., Hermsen, M., Manson, Q.F., Balkenhol, M.,
  et~al.:
\newblock Diagnostic assessment of deep learning algorithms for detection of
  lymph node metastases in women with breast cancer.
\newblock Jama \textbf{318} (2017)  2199--2210

\bibitem{gale2017detecting}
Gale, W., Oakden-Rayner, L., Carneiro, G., Bradley, A.P., Palmer, L.J.:
\newblock Detecting hip fractures with radiologist-level performance using deep
  neural networks.
\newblock arXiv preprint arXiv:1711.06504 (2017)

\bibitem{wang2017chestx}
Wang, X., Peng, Y., Lu, L., Lu, Z., Bagheri, M., Summers, R.M.:
\newblock Chestx-ray8: Hospital-scale chest x-ray database and benchmarks on
  weakly-supervised classification and localization of common thorax diseases.
\newblock In: 2017 IEEE Conference on Computer Vision and Pattern Recognition
  (CVPR), IEEE (2017)  3462--3471

\bibitem{staal2004ridge}
Staal, J., Abr{\`a}moff, M.D., Niemeijer, M., Viergever, M.A., Van~Ginneken,
  B.:
\newblock Ridge-based vessel segmentation in color images of the retina.
\newblock TMI \textbf{23} (2004)  501--509

\bibitem{gertych2007bone}
Gertych, A., Zhang, A., Sayre, J., Pospiech-Kurkowska, S., Huang, H.:
\newblock Bone age assessment of children using a digital hand atlas.
\newblock Computerized Medical Imaging and Graphics \textbf{31} (2007)
  322--331

\bibitem{rajpurkar2017mura}
Rajpurkar, P., Irvin, J., Bagul, A., Ding, D., Duan, T., Mehta, H., Yang, B.,
  Zhu, K., Laird, D., Ball, R.L.,  et~al.:
\newblock Mura dataset: Towards radiologist-level abnormality detection in
  musculoskeletal radiographs.
\newblock arXiv preprint arXiv:1712.06957 (2017)

\bibitem{heath2000digital}
Heath, M., Bowyer, K., Kopans, D., Moore, R., Kegelmeyer, P.:
\newblock The digital database for screening mammography.
\newblock Digital mammography (2000)  431--434

\bibitem{costa2005classification}
Costa, J.A., Hero, A.:
\newblock Classification constrained dimensionality reduction.
\newblock In: Acoustics, Speech, and Signal Processing, 2005.
  Proceedings.(ICASSP'05). IEEE International Conference on. Volume~5., IEEE
  (2005)  v--1077

\bibitem{fukunaga2013introduction}
Fukunaga, K.:
\newblock Introduction to statistical pattern recognition.
\newblock Academic press (2013)

\bibitem{jimenez1998supervised}
Jimenez, L.O., Landgrebe, D.A.:
\newblock Supervised classification in high-dimensional space: geometrical,
  statistical, and asymptotical properties of multivariate data.
\newblock IEEE Transactions on Systems, Man, and Cybernetics, Part C
  (Applications and Reviews) \textbf{28} (1998)  39--54

\bibitem{yang2016stacked}
Yang, Z., He, X., Gao, J., Deng, L., Smola, A.:
\newblock Stacked attention networks for image question answering.
\newblock In: Proceedings of the IEEE Conference on Computer Vision and Pattern
  Recognition. (2016)  21--29

\bibitem{trier1996feature}
Trier, {\O}.D., Jain, A.K., Taxt, T.:
\newblock Feature extraction methods for character recognition-a survey.
\newblock Pattern recognition \textbf{29} (1996)  641--662

\bibitem{srihari1999information}
Srihari, R., Li, W.:
\newblock Information extraction supported question answering.
\newblock Technical report, CYMFONY NET INC WILLIAMSVILLE NY (1999)

\bibitem{somers1999example}
Somers, H.:
\newblock Example-based machine translation.
\newblock Machine Translation \textbf{14} (1999)  113--157

\bibitem{bahdanau2014neural}
Bahdanau, D., Cho, K., Bengio, Y.:
\newblock Neural machine translation by jointly learning to align and
  translate.
\newblock arXiv preprint arXiv:1409.0473 (2014)

\bibitem{ciresan2012deep}
Ciresan, D., Giusti, A., Gambardella, L.M., Schmidhuber, J.:
\newblock Deep neural networks segment neuronal membranes in electron
  microscopy images.
\newblock In: Advances in neural information processing systems. (2012)
  2843--2851

\bibitem{hariharan2015hypercolumns}
Hariharan, B., Arbel{\'a}ez, P., Girshick, R., Malik, J.:
\newblock Hypercolumns for object segmentation and fine-grained localization.
\newblock In: Proceedings of the IEEE conference on computer vision and pattern
  recognition. (2015)  447--456

\bibitem{seyedhosseini2013image}
Seyedhosseini, M., Sajjadi, M., Tasdizen, T.:
\newblock Image segmentation with cascaded hierarchical models and logistic
  disjunctive normal networks.
\newblock In: Computer Vision (ICCV), 2013 IEEE International Conference on,
  IEEE (2013)  2168--2175

\bibitem{ronneberger2015u}
Ronneberger, O., Fischer, P., Brox, T.:
\newblock U-net: Convolutional networks for biomedical image segmentation.
\newblock In: International Conference on MICCAI, Springer (2015)  234--241

\bibitem{long2015fully}
Long, J., Shelhamer, E., Darrell, T.:
\newblock Fully convolutional networks for semantic segmentation.
\newblock In: Proceedings of the IEEE conference on computer vision and pattern
  recognition. (2015)  3431--3440

\bibitem{simonyan2014very}
Simonyan, K., Zisserman, A.:
\newblock Very deep convolutional networks for large-scale image recognition.
\newblock arXiv:1409.1556 (2014)

\bibitem{khurana2007comprehensive}
Khurana, A.:
\newblock Comprehensive ophthalmology.
\newblock New Age International Ltd (2007)

\bibitem{rezaee2017optimized}
Rezaee, K., Haddadnia, J., Tashk, A.:
\newblock Optimized clinical segmentation of retinal blood vessels by using
  combination of adaptive filtering, fuzzy entropy and skeletonization.
\newblock Applied Soft Computing \textbf{52} (2017)  937--951

\bibitem{turk1991face}
Turk, M.A., Pentland, A.P.:
\newblock Face recognition using eigenfaces.
\newblock In: Computer Vision and Pattern Recognition, 1991. Proceedings
  CVPR'91., IEEE Computer Society Conference on, IEEE (1991)  586--591

\bibitem{lyons1999automatic}
Lyons, M.J., Budynek, J., Akamatsu, S.:
\newblock Automatic classification of single facial images.
\newblock IEEE Transactions on pattern analysis and machine intelligence
  \textbf{21} (1999)  1357--1362

\bibitem{wu2002face}
Wu, J., Zhou, Z.H.:
\newblock Face recognition with one training image per person.
\newblock Pattern Recognition Letters \textbf{23} (2002)  1711--1719

\bibitem{moghaddam1998beyond}
Moghaddam, B., Wahid, W., Pentland, A.:
\newblock Beyond eigenfaces: Probabilistic matching for face recognition.
\newblock In: Automatic Face and Gesture Recognition, 1998. Proceedings. Third
  IEEE International Conference on, IEEE (1998)  30--35

\bibitem{akram2011retinal}
Akram, M.U., Tariq, A., Khan, S.A.:
\newblock Retinal recognition: Personal identification using blood vessels.
\newblock In: Internet Technology and Secured Transactions (ICITST), 2011
  International Conference for, IEEE (2011)  180--184

\bibitem{kuo2014kernel}
Kuo, B.C., Ho, H.H., Li, C.H., Hung, C.C., Taur, J.S.:
\newblock A kernel-based feature selection method for svm with rbf kernel for
  hyperspectral image classification.
\newblock IEEE Journal of Selected Topics in Applied Earth Observations and
  Remote Sensing \textbf{7} (2014)  317--326

\bibitem{crick2003textbook}
Crick, R.P., Khaw, P.T.:
\newblock A textbook of clinical ophthalmology: a practical guide to disorders
  of the eyes and their management.
\newblock (World Scientific)

\bibitem{akram2005common}
Akram, I., Rubinstein, A.:
\newblock Common retinal signs. an overview.
\newblock Optometry Today (2005)

\bibitem{tang2015contrast}
Tang, S., Huang, L., Wang, Y., Wang, Y.:
\newblock Contrast-enhanced ultrasonography diagnosis of fundal localized type
  of gallbladder adenomyomatosis.
\newblock BMC gastroenterology \textbf{15} (2015) ~99

\bibitem{noyel2017superimposition}
Noyel, G., Thomas, R., Bhakta, G., Crowder, A., Owens, D., Boyle, P.:
\newblock Superimposition of eye fundus images for longitudinal analysis from
  large public health databases.
\newblock Biomedical Physics \& Engineering Express \textbf{3} (2017)  045015

\bibitem{rib}
:
\newblock {Retina Image Bank: A project from the American Society of Retina
  Specialists}.
\newblock (\url{http://imagebank.asrs.org/about}) Accessed: 2018-06-30.

\bibitem{krizhevsky2012imagenet}
Krizhevsky, A., Sutskever, I., Hinton, G.E.:
\newblock Imagenet classification with deep convolutional neural networks.
\newblock In: Advances in NIPS. (2012)  1097--1105

\bibitem{xie2017aggregated}
Xie, S., Girshick, R., Doll{\'a}r, P., Tu, Z., He, K.:
\newblock Aggregated residual transformations for deep neural networks.
\newblock In: CVPR, IEEE (2017)  5987--5995

\bibitem{nair2010rectified}
Nair, V., Hinton, G.E.:
\newblock Rectified linear units improve restricted boltzmann machines.
\newblock In: Proceedings of the 27th international conference on machine
  learning (ICML-10). (2010)  807--814

\bibitem{hedeep}
He, K., Zhang, X., Ren, S., Sun, J.:
\newblock
\newblock (Deep residual learning for image recognition)

\bibitem{iandola2016squeezenet}
Iandola, F.N., Han, S., Moskewicz, M.W., Ashraf, K., Dally, W.J., Keutzer, K.:
\newblock Squeezenet: Alexnet-level accuracy with 50x fewer parameters and< 0.5
  mb model size.
\newblock arXiv:1602.07360 (2016)

\bibitem{yosinski2014transferable}
Yosinski, J., Clune, J., Bengio, Y., Lipson, H.:
\newblock How transferable are features in deep neural networks?
\newblock In: NIPS. (2014)  3320--3328

\bibitem{yang2018novel}
Yang, C.H.H., Huang, J.H., Liu, F., Chiu, F.Y., Gao, M., Lyu, W., Tegner, J.,
  et~al.:
\newblock A novel hybrid machine learning model for auto-classification of
  retinal diseases.
\newblock arXiv preprint arXiv:1806.06423 (2018)

\bibitem{chang2011libsvm}
Chang, C.C., Lin, C.J.:
\newblock Libsvm: a library for support vector machines.
\newblock TIST \textbf{2} (2011) ~27

\bibitem{zhou2015cnnlocalization}
Zhou, B., Khosla, A., A., L., Oliva, A., Torralba, A.:
\newblock {Learning Deep Features for Discriminative Localization.}
\newblock CVPR (2016)

\end{thebibliography}

\end{document}